# Symptom based Hierarchical Classification of Diabetes and Thyroid disorders using Fuzzy Cognitive Maps

Anand. M. Shukla[1], Pooja. D. Pandit[1],
Vasudev. M. Purandare[1].
Student

Anuradha Srinivasaraghavan[1]
Associate Professor,
Research Scholar

[1]St. Francis Institute of Technology, Borivali (West).
Maharashtra, India

*Abstract*—Fuzzy Cognitive Maps (FCMs) are soft computing technique that follows an approach similar to human reasoning and human decision-making process, making them a valuable modeling and simulation methodology. Medical Decision Systems are complex systems consisting of many factors that may be complementary, contradictory, and competitive; these factors influence each other and determine the overall diagnosis with a different degree. Thus, FCMs are suitable to model Medical Decision Support Systems. The proposed work therefore uses FCMs arranged in hierarchical structure to classify between Diabetes, Thyroid disorders and their subtypes. Subtypes include type 1 and type 2 for diabetes and hyperthyroidism and hypothyroidism for thyroid.

*Keywords—Fuzzy Cognitive Maps(FCM), Nonlinear Hebbian Learning algorithm(NHL), Aritifical Neural Netwroks(ANN).*

## I. INTRODUCTION

Fuzzy Cognitive Maps (FCMs) were introduced by B. Kosko as a knowledge-based methodology for modeling and simulating dynamic systems [1]. FCMs are similar to Artificial Neural Networks (ANN), but FCM has both inter-layered and intra-layered connections amongst the neurons. FCM can be represented as a directed graph, where nodes are the concepts and the edges represent weighted links among them. Concepts represent variables, inputs and outputs, which are essential to model a system. The concepts are connected by links representing cause-effect relationships among them. Each link is associated with a weight that signifies the degree to which each concept affects the other. Any of the concepts of FCM can be defined either as input concepts or output concepts (OCs). Output concepts are those factors which characterize the system and are of interest i.e. their estimated values represent the final state of the system [1].

FCM are designed based on the inputs obtained from group of experts who define the concepts and describe the relationships among those concepts. Experts help determine weights Wij for all edges and the weights lie in the range of -1 to 1.

Thus an initial matrix

$W_{initial}$ = [$w_{ji}$] , i,j=1,……,N and, i=1,..,N,  is obtained.

Each concept is associated with a number $A_i$ that represents its value and it lies in the interval [0,1]. The interconnection strength between two nodes $C_j$ and $C_i$ is $w_{ji}$, with $w_{ji}$ taking on any value in the range -1 to 1. There are three possible types of causal relationships among concepts:

- $W_{ji}$ is positive indicating direct proportionality
- $W_{ji}$ is negative indicating inverse proportionality
- $W_{ji}$ is zero indicating no relationship

Generally, the value of each concept is calculated, computing the influence of other concepts to the specific concept, by applying the following equation:

$$A_i^{(k+1)} = f\left(A_i^{(k)} + \sum_{\substack{j \neq i \\ j=1}}^{N} A_j^{(k)} \cdot w_{ji}^{(k)}\right) \quad (1)$$

where $A_i^{(k+1)}$ is the value of concept $C_i$ at time k +1, $A_j^{(k)}$ is the value of concept $C_j$ at time k, $w_{ji}$ is the weight of the interconnection between concept $C_j$ and concept $C_i$ and f is the sigmoid threshold function [2].

Given initial input concept vector $A_0$ the equation (1) is used iteratively to calculate values of output concepts, until the values do not change significantly. The final concept vector represents the steady state of FCM i.e. convergence. Experts involved in the construction of FCM determine concepts and causality among them. This approach may yield to a distorted model. Therefore, using initial weight matrix it is not always guaranteed that FCM will converge to a desired state. In order to achieve the desired convergence, it is necessary to fine-tune the initial weights for that learning algorithms need to be applied [1].

## II. NON-LINEAR HEBBIAN LEARNING RULE

The NHL algorithm is based on the premise that all the concepts in FCM model are triggering at each iteration step and change their values [1].

This simple rule states that if $A_i^{(k)}$ is the value of concept $C_i$ at iteration k , and $A_j$ is the value of the triggering concept $C_j$ which triggers the concept $C_i$, the corresponding weight $w_{ji}$ from concept $C_j$ towards the concept $C_i$ is increased proportional to their product multiplied with the learning rate parameter minus the weight decay at iteration step k [1].

The training weight algorithm takes the following form:

$$w_{ji}^{(k)} = \gamma \cdot w_{ji}^{(k-1)} + \eta \cdot A_i^{(k-1)}\left(A_j^{(k-1)} - sgn(w_{ji}^{(k-1)}) w_{ji}^{(k-1)} A_i^{(k-1)}\right) \quad (2)$$

where *y* is the weight decay learning coefficient which prevents values of weights from reaching 1 and coefficient $\eta_k$ is a very small positive scalar factor called learning parameter[1].

At every simulation step the value of each concept of FCM is updated, using the equation (1) whereas the value of weight $w_{ji}^{(k)}$ is calculated with equation (2).

The NHL algorithm does not assign new interconnections and all the zero weights do not change value. When the algorithm termination conditions are met, the final weight matrix $W_{NHL}$ is derived [1]. This termination criterion is based on the variation of the subsequent values of $DOC_j$ concepts, for iteration step k, yielding a very small value e, taking the form:

$$\left|OC_j^{(k+1)} - OC_j^{(k)}\right| < e$$

This criterion determines when the process of the learning terminates. The term e is a tolerance level keeping the variation of values of OC(s) as low as possible and it is proposed equal to e = 0.001.

### III. FCM AS CLASSIFIER

The values of output concepts at convergence represent a future steady state. This property of FCM to reach equilibrium by simulating the cause-effect relationships between the given input concepts can be used along with the fundamentals of competitive learning to model a classifier which convergences to a different region depending upon the given input. Competitive learning is a form of learning where output neurons compete against each other to produce the output and this winner is undergoes learning.

### IV. DEFINING DISORDER

Diabetes has become a common killer nowadays. The people in age group of 20 to 79 years are at risk. More than 44 lakh Indians in this age group are unaware that they are diabetic. Diabetes increases risks of other severe complications like heart attack, stroke, kidney disorder, vision problems and in extreme cases amputation. India is presently home to 62 million diabetics - an increase of nearly 2 million in just one year. India is second only to China which is home to 92.3 million diabetics. By 2030, India's diabetes numbers are expected to cross the 100 million mark.

Thyroid diseases are, arguably, among the commonest endocrine disorders worldwide. According to a projection from various studies on thyroid disease, it has been estimated that about 42 million people in India suffer from thyroid diseases.

There is therefore a significant need to reach out and make people aware of the causes, symptoms, treatment and importance of testing for thyroid and diabetes problems.

### V. PROPOSED CLASSIFIER

The proposed classifier consists of three FCMs arranged in a hierarchical manner. The first stage of classification is the distinction between diabetes and thyroid using FCM1 (shown in figure 2). This FCM1 gives an indication of whether the patient symptoms more closely resemble diabetes or thyroid case. The inference acquired from FCM1 is used to decide which FCM from level 2 will be used for further classification. This means that diabetes cases will ideal lead to selection of FCM2 (shown in figure 5). which then classifies the patient as Type1 or Type 2. Similarly a thyroid case will lead to selection of FCM3 (shown in figure 8). for further classification between hyperthyroidism and Hypothyroidism. The proposed classifier structure is given in figure (1) below:

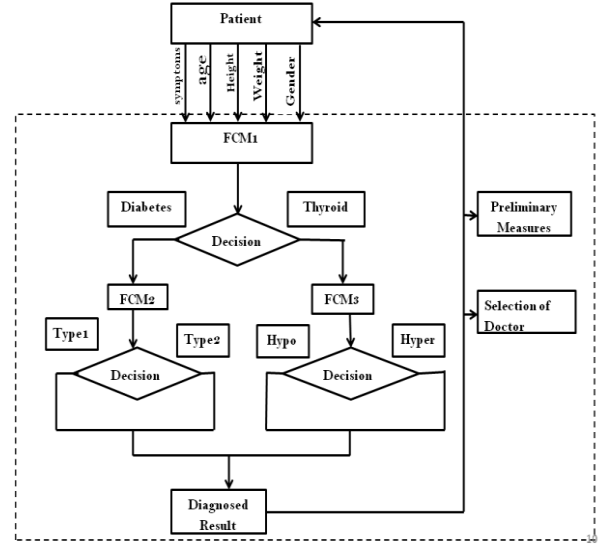

Fig. 1. Proposed FCM based classifier

### VI. CONSTRUCTING FCM

Both diabetes and thyroid disorders show some early symptoms. These symptoms can be used to suggest the chances of patient having either of the two disorders. Also there are some lifestyle factors and genetic factors that can help determine if a particular person is at risk of diabetes and thyroid. So using these symptoms and risk factors a FCM can be modeled to predict the chances of a person suffering from diabetes and thyroid based on the symptoms and their severities which are given as input by the patient.

The symptoms associated with diabetes include: Frequent urination, Frequent Thirst, Change in appetite (increased hunger), Blurred vision, etc.

The symptoms associated with thyroid include: Fatigue, Skin Problems, Weight variation (weight gain/loss), Irritability, Change in appetite, Nausea, Vomiting, Hair loss, etc.

From the above list, the common symptoms between diabetes and thyroid can be used as the input concepts for the FCM that classifies between the two.

FCM aims in classifying diabetes and thyroid. So the proposed model has these as its output concepts. Each input concept is connected to both the output concepts, but these edges are not causal links they represent association of symptoms with disease. The weights on these edges define the degree of association. The weights lie in the range of -1 to 1, and this range can be divided into five regions of memberships: Positively Strong, Positively weak, Neutral, Negatively weak, and Negatively Strong. These linguistic labels are used to get weights from the experts. After defuzzifying these labels, an initial weight matrix can be derived as shown in table 1. The initial FCM is derived is shown in figure (2).

## FCM FOR IDENTIFYING DIABETES AND THYROID

TABLE I
Concepts for FCM that classifies Diabetes and Thyroid

| | |
|---|---|
| C1 | Diabetes |
| C2 | Thyroid |
| C3 | Fatigue |
| C4 | Change in Appetite |
| C5 | Weight Variation |
| C6 | Vision Problems |
| C7 | Skin Problems |
| C8 | Irritability |
| C9 | Trembling |

TABLE II
Initial weight matrix for FCM that classifies Diabetes and Thyroid

| | C1 | C2 | C3 | C4 | C5 | C6 | C7 | C8 | C9 |
|---|---|---|---|---|---|---|---|---|---|
| C1 | 1.0 | -1.0 | 0 | 0 | 0 | 0 | 0 | 0 | 0 |
| C2 | -1.0 | 1.0 | 0 | 0 | 0 | 0 | 0 | 0 | 0 |
| C3 | 0.6 | 0.8 | 1.0 | 0 | 0.25 | 0 | 0 | 0.4 | 0.15 |
| C4 | 0.8 | 0.5 | 0.15 | 1.0 | 0.45 | 0 | 0 | 0.15 | 0 |
| C5 | 0.7 | 0.6 | 0.75 | 0.3 | 1.0 | 0 | 0 | 0 | 0 |
| C6 | 0.3 | 0.4 | 0 | 0 | 0 | 1.0 | 0 | 0 | 0 |
| C7 | 0.7 | 0.8 | 0 | 0 | 0 | 0 | 1.0 | 0 | 0 |
| C8 | 0.3 | 0.4 | 0.2 | 0.2 | 0.14 | 0 | 0 | 1.0 | 0.5 |
| C9 | 0.3 | 0.5 | 0 | 0 | 0 | 0 | 0 | 0 | 1.0 |

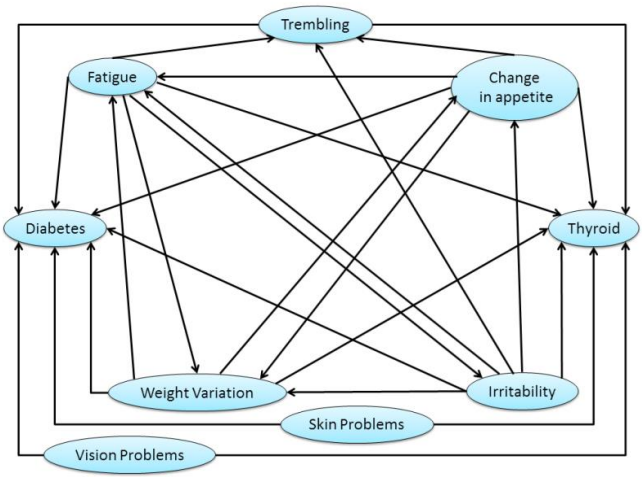

Fig. 2. FCM that classifies Diabetes and Thyroid (FCM1)

## VII. IMPLEMENTATION

Symptoms and their severities taken from experts are given as input to the FCM. Using these inputs and applying equation (1) on initial weight matrix, final converged values of output concepts are calculated. Comparing these values, the maximum of the two is considered as the result i.e. if the output concept value of diabetes is greater than thyroid then the given input is considered as diabetes case.

The vector below depicts the severity of the symptoms for ideal diabetes case.

A0 = {0.0  0.0  0.6  0.8  0.7  0.3  0.7  0.3  0.3}

Using equation (1) iteratively on A0 and initial weight matrix, the final converged concept vector is as follows:

{ 0.89578, 0.91107, 0.66187, 0.57555, 0.62422, 0.50000, 0.50000, 0.58707, 0.59704}

The value of output concept Thyroid is greater than Diabetes which indicates that the input concept vector is of Thyroid. But input vector provided was of Diabetes case. So FCM is not converging to the desired region and it needs training. Therefore, NHL algorithm is applied to make FCM converge into the desired region.

But the property of NHL algorithm states that after a final weight matrix is obtained any random input will converge to the same desired region [3]. Also the changes in weights are very drastic. This is not desirable for a classifier. So the FCM is to be trained separately to get two weight matrices that converge to two different convergence region i.e. one for diabetes and other for thyroid. After obtaining these two matrices the final matrix is average of these two.

Also when FCM is used to model a classifier, it has connections (arcs) between the output concepts. These are not cause–effect connections, but inhibitory connections. In the applications where FCMs are used to establish a diagnosis, the disorder concepts are considered outputs. In order to achieve the correct diagnosis with the highest probability, the output concepts must ''compete'' against each other for only one of them to dominate. To achieve this ''competition'' between output concepts, the interaction of each of the node with the others should have a very high negative weight (even -1).This implies that the higher value of a given node, should lead to a lowering of the value of competing nodes, i.e. strong inhibition [4]. So the initial weight matrix changes to include this negative feedback of -1 from C1 to C2 and C2 to C1.

A modified Case Based FCM reasoning formula can be used to remove the spurious influence of inactive concepts (concepts with zero values) on other concepts, to avoid the emergence of conflicts in cases where the initial values of concepts are 0.5, and to overcome missing data [5]. We therefore reformulated Eq. (1) as:

$$C_i^{(k+1)} = f\left(\left(2C_i^{(k)} - 1\right) + \sum_{\substack{j \neq i \\ j=i}}^{N}\left(2C_j^{(k)} - 1\right) \cdot W_{ji}\right) \quad (3)$$

Finally now using equation (3) with NHL algorithm final matrix of FCM to classify diabetes and thyroid is:

{1.00000, -1.00000, 0.00000, 0.00000, 0.00000, 0.00000, 0.00000, 0.00000, 0.00000},

{-1.00000, 1.00000, 0.00000, 0.00000, 0.00000, 0.00000, 0.00000, 0.00000, 0.00000},

{0.52788, 0.70129, 1.00000, 0.00000, 0.22273, 0.00000, 0.00000, 0.35352, 0.13545},

{0.70197, 0.44037, 0.13540, 1.00000, 0.39698, 0.00000, 0.00000, 0.13532, 0.00000},

{0.61493, 0.52737, 0.65864, 0.26621, 1.00000, 0.00000, 0.00000, 0.00000, 0.00000},

{0.26517, 0.35215, 0.00000, 0.00000, 0.00000, 1.00000, 0.00000, 0.00000, 0.00000},

{0.61754, 0.70438, 0.00000, 0.00000, 0.00000, 0.00000, 1.00000, 0.00000, 0.00000},

{0.26631, 0.35350, 0.17903, 0.17894, 0.12665, 0.00000, 0.00000, 1.00000, 0.44061},

{0.26627, 0.44038, 0.00000, 0.00000, 0.00000, 0.00000, 0.00000, 0.00000, 1.00000},

Now using this weight matrix and equation (3): The final converged vector for diabetes input is:

{ 0.56406 0.51207 0.50291 0.50136 0.50215 0.30000 0.70000 0.50183 0.50182 }

Here clearly the value of diabetes is more which should be the proper output. This convergence is graphical represented by figure (3). The graph clearly shows that the values of output concept stabilize after some 'n' number of iterations. And it can also be seen that output value of diabetes is more suggesting that input given is a diabetes case, which is true.

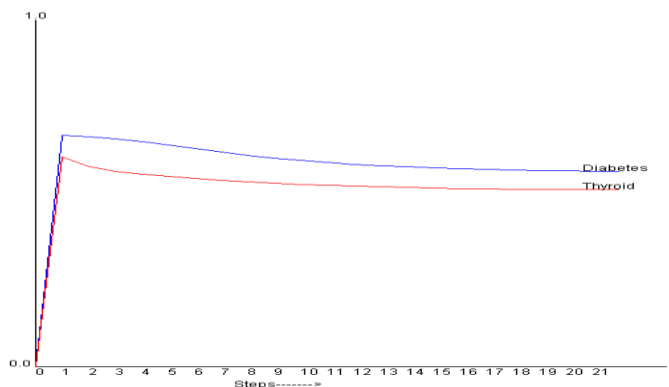

Fig. 3. Diabetes input given to FCM1

And the final converged vector for thyroid input is:

{ 0.49176 0.66919 0.50023 0.50011 0.50017 0.40000 0.80000 0.50014 0.50014 }
The graphical representation of same is shown is figure (4).

Not only the major illness, but also the major sub categories of these illnesses are categorized. So a FCM is constructed which helps in diagnosing between hyperthyroidism and hypothyroidism in the case of thyroid or type 1 and type 2 in the case of diabetes. The input concepts are again the symptoms and disorders are the output. The same approach (NHL) is used to get final weight matrix for FCM that classifies between Type1 and Type2 and FCM that classifies between Hyperthyroidism and Hypothyroidism, which are then used for classification.

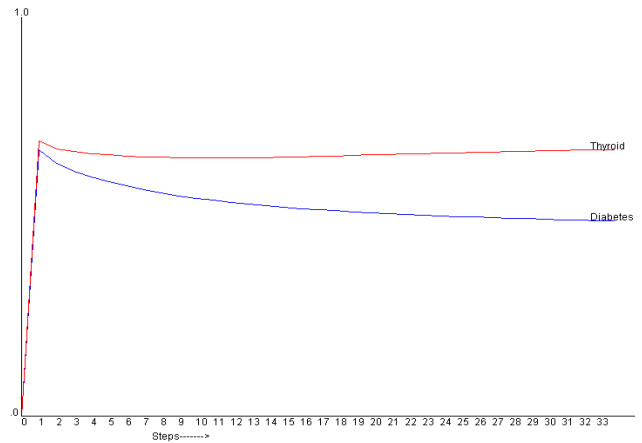

Fig. 4. Thyroid input given to FCM1

## FCM FOR IDENTIFYING TYPE1 AND TPYE 2 DIAETES

TABLE IV
Concepts for FCM that classifies Diabetes Type1 and Type 2

| C1 | Type 1 |
|----|--------|
| C2 | Type 2 |
| C3 | Frequent Urination |
| C4 | Frequent Thirst |
| C5 | Nausea |
| C6 | Vomiting |
| C7 | Gum Problems |
| C8 | Erectile Dysfunction (ED) |

TABLE V
Initial weight matrix for FCM that classifies Diabetes Type1 and Type 2

|    | C1   | C2   | C3   | C4    | C5   | C6   | C7  | C8  |
|----|------|------|------|-------|------|------|-----|-----|
| C1 | 1.0  | -1.0 | 0.0  | 0.0   | 0.0  | 0.0  | 0.0 | 0.0 |
| C2 | -1.0 | 0.0  | 0.0  | 0.0   | 0.0  | 0.0  | 0.0 | 0.0 |
| C3 | 0.3  | 0.7  | 1.0  | 0.5   | 0    | 0    | 0   | 0   |
| C4 | 0.7  | 0.8  | 0.75 | 1.0   | 0.25 | 0.25 | 0   | 0   |
| C5 | 0.5  | 0    | 0    | -0.15 | 1.0  | 0.8  | 0   | 0   |
| C6 | 0.6  | 0.7  | -0.5 | 0.3   | 0.5  | 1.0  | 0   | 0   |
| C7 | 0    | 0.7  | 0    | 0     | 0    | 0    | 1.0 | 0   |
| C8 | 0.1  | 0.6  | 0    | 0     | 0    | 0    | 0   | 1.0 |

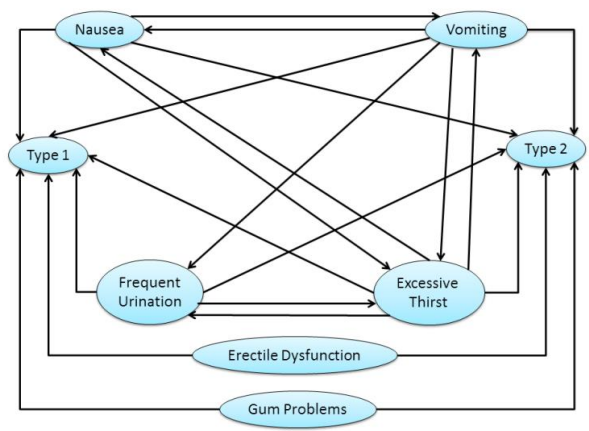

Fig. 5. FCM that classifies Diabetes Type1 and Type 2 (FCM2)

Applying NHL along with equation (3) it is possible to obtain a final weight matrix for FCM that classifies diabetes Type1 and Type2 by following the same approach of training used for previous FCM.

The Final weight matrix for this FCM is given as follows:

{1.00000, -1.00000, 0.00000, 0.00000, 0.00000, 0.00000, 0.00000, 0.00000},

{-1.00000, 1.00000, 0.00000, 0.00000, 0.00000, 0.00000, 0.00000, 0.00000},

{0.78338, 0.68548, 1.00000, 0.48992, 0.00000, 0.00000, 0.00000, 0.00000},

{0.68554, 0.78330, 0.73457, 1.00000, 0.24524, 0.24524, 0.00000, 0.00000},

{0.48981, 0.00000, 0.00000, -0.14640, 1.00000, 0.78347, 0.00000, 0.00000},

{0.58767, 0.39187, -0.48934, 0.29415, 0.48988, 1.00000, 0.00000, 0.00000},

{0.00000, 0.68556, 0.00000, 0.00000, 0.00000, 0.00000, 1.00000, 0.00000},

{0.09813, 0.58763, 0.00000, 0.00000, 0.00000, 0.00000, 0.00000, 1.00000},

## FCM FOR IDENTIFYING HYPERTHYROIDISM and HYPOTHYROIDISM

TABLE VI
Concepts for FCM that classifies thyroid subtypes

| C1 | Hyperthyroidism |
| C2 | Hypothyroidism |
| C3 | Hair Loss |
| C4 | Heart Rate |
| C5 | Heat / Cold tolerance |
| C6 | Constipation |
| C7 | Diarrhea |
| C8 | Mental Problems |
| C9 | Menstrual Problems |
| C10 | Breathlessness |

TABLE V
Initial weight matrix for FCM that classifies Thyroid subtypes

|  | C1 | C2 | C3 | C4 | C5 | C6 | C7 | C8 | C9 | C10 |
| --- | --- | --- | --- | --- | --- | --- | --- | --- | --- | --- |
| C1 | 1 | -1.0 | 0 | 0 | 0 | 0 | 0 | 0 | 0 | 0 |
| C2 | -1.0 | 1 | 0 | 0 | 0 | 0 | 0 | 0 | 0 | 0 |
| C3 | 0.7 | 0.8 | 1 | 0 | 0 | 0 | 0 | 0 | 0 | 0 |
| C4 | 0.9 | 0.1 | 0 | 1 | 0 | 0 | 0.5 | 0.5 | 0 | 0.8 |
| C5 | 0.8 | 0.89 | 0 | 0.5 | 1 | 0 | 0 | 0 | 0 | 0.2 |
| C6 | 0.1 | 0.6 | 0 | 0 | 0 | 1 | -0.7 | 0 | 0 | 0.1 |
| C7 | 0.4 | 0 | 0.6 | 0.6 | 0 | -0.7 | 1 | 0 | 0 | 0.1 |
| C8 | 0.8 | 0.6 | 0 | 0 | 0 | 0 | 0 | 1 | 0 | 0 |
| C9 | 0.8 | 0.65 | 0.5 | 0.5 | 0 | 0.4 | 0.4 | 0 | 1 | 0.15 |
| C10 | 0.6 | 0.3 | 0.5 | 0.7 | 0 | 0.5 | 0 | 0 | 0 | 1.0 |

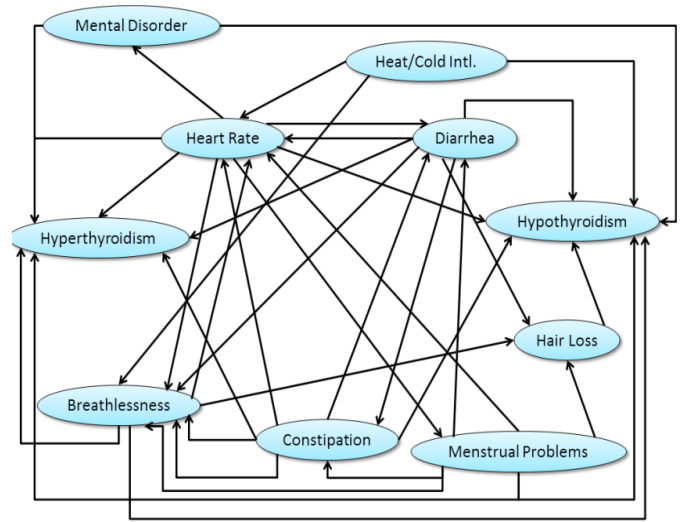

Fig. 6. FCM that classifies thyroid subtypes (FCM3)

Applying NHL along with equation (3) it is possible to obtain a final weight matrix for FCM that classifies Hyperthyroidism and Hypothyroidism. The Final weight matrix for this FCM is given as follows:

{1.00000, -1.00000, 0.00000, 0.00000, 0.00000, 0.00000, 0.00000, 0.00000, 0.00000, 0.00000},

{-1.00000, 1.00000, 0.00000, 0.00000, 0.00000, 0.00000, 0.00000, 0.00000, 0.00000, 0.00000},

{0.54915, 0.94061, 1.00000, 0.00000, 0.00000, 0.00000, 0.00000, 0.00000, 0.00000, 0.00000},

{0.76999, 0.44610, 0.00000, 1.00000, 0.00000, 0.00000, 0.53060, 0.52982, 0.00000, 0.75725},

{0.74610, 0.95426, 0.00000, 0.52839, 1.00000, 0.00000, 0.00000, 0.00000, 0.00000, 0.30231},

{0.10296, 0.74687, 0.00000, 0.00000, 0.00000, 1.00000, -0.58836, 0.00000, 0.00000, 0.19869},

{0.38751, 0.04500, 0.57635, 0.57329, 0.00000, -0.59076, 1.00000, 0.00000, 0.00000, 0.19680},

{0.71594, 0.63869, 0.00000, 0.00000, 0.00000, 0.00000, 0.00000, 1.00000, 0.00000, 0.00000},

{0.78953, 0.77810, 0.52446, 0.52148, 0.00000, 0.44873, 0.44341, 0.00000, 1.00000, 0.25750},

{0.61002, 0.36152, 0.52390, 0.67270, 0.00000, 0.51977, 0.00000, 0.00000, 0.00000, 1.00000},

## VIII. RESULTS

The accuracy of FCM1 is tested on 22 input concept vectors. This input sample contains equal numbers of diabetes and thyroid cases. The accuracy is measured using the following confusion matrix:

|          | Diabetes | Thyroid |
|----------|----------|---------|
| Diabetes | 10       | 1       |
| Thyroid  | 3        | 8       |

Accuracy =18/22= 81.8182 %   Error =4/22=18.1818 %

The accuracy of FCM2 is tested on 18 input concept vectors. This input sample contains different numbers of diabetes type 1 and type 2 cases. The accuracy is measured using the following confusion matrix:

|        | Type 1 | Type 2 |
|--------|--------|--------|
| Type 1 | 9      | 0      |
| Type 2 | 3      | 6      |

Accuracy =15/18= 83.333 %   Error =3/18=16.6667 %

The accuracy of FCM3 is tested on 24 input concept vectors. This input sample contains different numbers of Hyperthyroidism and Hypothyroidism cases. The accuracy is measured using the following confusion matrix:

|                 | Hyperthyroidism | Hypothyroidism |
|-----------------|-----------------|----------------|
| Hyperthyroidism | 9               | 2              |
| Hypothyroidism  | 8               | 5              |

Accuracy =15/18= 58.3333 %   Error =3/18=41.6667 %

The accuracy of FCM3 is poor due to the fact that it converges all inputs to almost equal output concept values. This flaw remains true even if NHL is applied repeatedly on the initial weight matrix. In this FCM the initial matrix is assumed to be distorted or incomplete. But in spite of faulty convergence this FCM show correct inference if the path followed to convergence is observed.

## IX. RELATED WORK

The concept of Fuzzy Cognitive Maps (FCM) was introduced by [1] with a focus on modeling and simulating dynamic systems. This was further developed by novel and excellent techniques proposed in [2], [3], [4], [5], [6] and [7].

There is a lot of work on analytical modeling based approaches in literature such as [8], [9], [10], [11], [12], [13], [14], [15] which take a mathematical approach to problem solving. In contrast, the focus of our work is orthogonal to such approaches as we take a learning based route.

Miniaturized antennas [16], [17], [18], [19], [20], [21], [22], [23], [24], [25], [26], [27], [28], [29], [30] have been used for detection of medical disorders. With the help of antenna implantation in a patient, such work collect data which aids in detection and diagnosis. In contrast, our work is passive and does not require any kind of active implantation or surgeries.

## X. CONCLUSION

The proposed hierarchical classifier therefore takes symptoms as input from the patient and presents a final diagnosis to the patient. The classifier accuracy depends upon accuracy of individual FCMs. In order to improve accuracy of the proposed classifier it is necessary to consider some additional facts regarding the disorders, these facts mostly include risk factors and statistics. For e.g. it is known that females are more likely to develop thyroid disorders, hypothyroidism cases are more common than hyperthyroidism, also type 2 diabetes is more common than type1, and many more. Risk factors need to be considered especially in case of diabetes because it is a lifestyle based disorder. Also genetic factors can help in better classification as they affect the chances of person suffering from a particular disorder. This additional information helps in making the FCM more biased towards a particular disorder, for a specific input. Making FCM biased can be as simple as setting output concept value of the particular disorder higher than the other, or suppressing the inhibitory feedback from the other output concepts to this disorder output. In order to do this it is necessary to take the risk factors input from user and set the output concept values accordingly before simulation of input on FCM.

ACKNOWLEDGMENT

We wish to offer our sincere thanks to each and every person who has helped us either directly or indirectly during the course of this work. We would like to express our deep gratitude towards our guide Ms. Anuradha Srinivasaraghavan for her constant guidance, support for sharing her experience and knowledge and pushing us in the right direction to ensure timely completion of all our work. Special thanks to Dr. Hafeezunisa for providing us required information about the disorders.